\pdfoutput=1

\documentclass[11pt]{article}

\usepackage[table]{xcolor}
\usepackage[final]{acl}

\usepackage{times}
\usepackage{latexsym}

\usepackage[T1]{fontenc}

\usepackage[utf8]{inputenc}

\usepackage{microtype}

\usepackage{inconsolata}

\usepackage{graphicx}
\usepackage{paralist}
\usepackage{algorithm}
\usepackage{algorithmic}
\usepackage{amsmath}
\usepackage{multirow}
\usepackage{booktabs}
\usepackage{colortbl}

\title{\raisebox{-0.4em}{\includegraphics[height=1.5em]{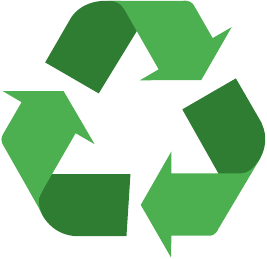}}\ Turning Trash into Treasure: Accelerating Inference of Large Language Models with Token Recycling}

\author{
    Xianzhen Luo\textsuperscript{\rm 1}, Yixuan Wang\textsuperscript{\rm 1}, Qingfu Zhu\textsuperscript{\rm 1}\thanks{Corresponding author}, Zhiming Zhang\textsuperscript{\rm 1}, \\ 
    \textbf{Xuanyu Zhang}\textsuperscript{\rm 2},\textbf{Qing Yang}\textsuperscript{\rm 2}, \textbf{Dongliang Xu}\textsuperscript{\rm 2}  \\
    \textsuperscript{\rm 1}Harbin Institute of Technology, Harbin, China\\
    \textsuperscript{\rm 2}Du Xiaoman (Beijing) Science Technology Co., Ltd. \\
    \texttt{\{xzluo, wyx, qfzhu, zmzhang\}@ir.hit.edu.cn} \\
    \texttt{\{zhangxuanyu, yangqing, xudongliang\}@duxiaoman.com} \\
}

\begin{document}
\maketitle
\begin{abstract}
Massive parameters of LLMs have made inference latency a fundamental bottleneck.
Speculative decoding represents a lossless approach to accelerate inference through a guess-and-verify paradigm.
Some methods rely on additional architectures to guess draft tokens, which need extra training before use. 
Alternatively, retrieval-based training-free techniques build libraries from pre-existing corpora or by n-gram generation. However, they face challenges like large storage requirements, time-consuming retrieval, and limited adaptability.
Observing that candidate tokens generated during the decoding process are likely to reoccur in future sequences, we propose Token Recycling. 
It stores candidate tokens in an adjacency matrix and employs a breadth-first-search (BFS)-like algorithm to construct a draft tree, which is then validated through tree attention. 
New candidate tokens from the decoding process are then used to update the matrix.
Token Recycling requires \textless2MB of additional storage and achieves approximately 2x speedup across all sizes of LLMs. It significantly outperforms existing train-free methods by 30\% and even a widely recognized training method by 25\%. 
\end{abstract}

\begin{figure}[!ht]
\centering
\includegraphics[width=\columnwidth]{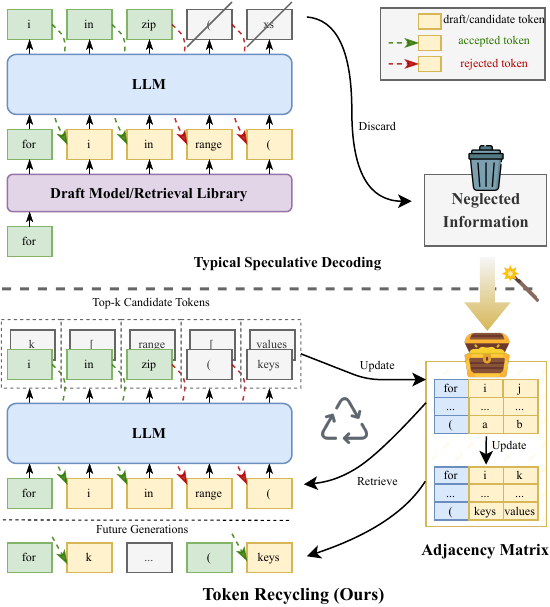} 
\caption{A comparison of typical speculative decoding and Token Recycling (TR). 
Typical methods draft some tokens and verify them in parallel in one decoding step. 
Unlike other methods that discard candidate tokens, TR stores them in an adjacency matrix. In future generations, draft tokens are retrieved from the matrix which is updated with new candidate tokens.
TR effectively recycles tokens in the decoding process.
}
\label{fig:intro}
\end{figure}

\section{Introduction}


Large Language Models (LLMs) \cite{gpt3,team2023gemini,llama2,llama3} have 
becoming the foundation of numerous applications such as chatbots, code assistants, and agents~\cite{openai2023gpt4,chen2021EvaluatingLargeLanguage,wang2024Surveya}.
However, due to the \textit{auto-regressive} decoding strategy, LLMs can only generate a single token at each decoding step, leading to high inference latency~\cite{gpt3}.
The latency mainly comes from transferring billions of parameters from high bandwidth memory to the accelerator cache at each decoding step, rather than arithmetic computations~\cite{kim2024SqueezeLLM,shazeer2019Fast,cai2024MedusaSimpleLLM}. 

Many approaches~\cite{xu2024onebit,pmlr-v202-frantar23a,dao2023flashattention2, deepseekv2} seek to reduce the latency, with \textit{speculative decoding} as a key lossless technique. 
This approach employs a \textit{guess and verify} process to obtain multiple tokens during a single decoding step~\cite{chen2023Accelerating,leviathan2023Fast, miao2024SpecInfer, xia-etal-2023-speculative}. 
It first speculates several subsequent draft tokens and then verifies them using the original LLMs. 
The time cost of verification on multiple tokens is comparable to that of generating one token due to the high parallelism of accelerators.
Once some draft tokens are correct, the decoding steps is significantly shortened without sacrificing quality.
To fully utilize the parallelism of accelerators, \textit{tree attention} slightly adjust the attention mask to verify multiple token sequences in one model forward~\cite{cai2024MedusaSimpleLLM, miao2024SpecInfer}.

For effective acceleration, speculative decoding must ensure accurate draft predictions while keeping speculation overhead low.
Additional model architectures are constructed to guess the draft tokens, 
including small draft models~\cite{leviathan2023Fast,chen2023Accelerating} and parameter-efficient structures~\cite{cai2024MedusaSimpleLLM, lin2024BiTA}.
However, these approaches require resources for additional training on each LLM.
The typical approach to achieve train-free speculative decoding is retrieve-based. 
In this case, a retrieval library is pre-defined to obtain tokens following the suffix of current content as draft tokens.
Several methods have been proposed in this category, each with its trade-offs:
\begin{inparaenum}[\bfseries (i)]
\item REST~\cite{he2023rest} transforms existing corpora into a retrieval library, but \textit{the storage is large, retrieval is time-consuming}, and \textit{the library lacks flexibility} as it's static to any queries.
\item PLD~\cite{saxena2023prompt} only retrieves the previous content with minimal cost. However, \textit{it can not predict new tokens or new token combinations}.
\item Lookhead~\cite{fu2024BreakSequentialDependency} construct and update an n-gram library by decoding n times with LLMs. However, \textit{LLMs have to generate n-grams while in inference, causing low efficiency}. 
\end{inparaenum}

Furthermore, \textbf{all speculative decoding approaches fail to fully utilize candidate tokens}, which are multiple possible next tokens generated by LLMs at each decoding step. 
In greedy decoding, only the top-1 candidate token of accepted tokens is selected as the output, while other candidate tokens, 
including all candidate tokens from rejected tokens, 
are discarded, such as `k' and `keys' in Figure~\ref{fig:intro}.
However, we observe that \textbf{when current input tokens reappear in future generations, the following tokens could be candidate tokens generated several steps prior}. 
Based on the observation, we propose \textbf{Token Recycling (TR)}, which utilizes candidate tokens as draft tokens.
It stores candidate tokens in an adjacency matrix.
Before each decoding step, 
a BFS-like approach retrieves a draft tree from the matrix, which is then verified using tree attention. 
Once verified, the newly generated candidate tokens update the matrix.
\begin{inparaenum}[\bfseries (i)]
\item The matrix provides a \textbf{flexible retrieval library} that is tailored to each query and offers \textbf{low retrieval costs} due to its \textbf{small size (\textless2MB)}. 
\item Compared to using the previous content solely, candidate tokens naturally include more tokens, providing \textbf{many possible continuations}. 
\item The construction and update of our library (matrix) utilize the `trash' tokens \textbf{without requiring any additional generation}. 
\end{inparaenum}

We conduct comprehensive experiments on general benchmark SpecBench~\cite{xia2024UnlockingEfficiencyLarge}, and specialized dataset on code domain, MBPP ~\cite{chen2021EvaluatingLargeLanguage} with Vicuna~\cite{zheng2023JudgingLLMasaJudgeMTBench} and Code Llama~\cite{roziere2023code} . The results show that TR greatly exceeds previous train-free approaches, and improves more than 30\% on all sizes (7b, 13b, 33b/34b).
The speed-up ratio even exceeds the widely used training approach--Medusa, demonstrating its high efficiency.

Our contributions are summarized below:
\begin{itemize}
    \item A plug-and-play speculative decoding method, Token Recycling is proposed. It firstly recognizes the value of `trash' tokens and converts them into `treasure' tokens for acceleration.
    \item TR requires minimal storage space (\textless2MB) with a low retrieval cost and covers many new tokens. Continuously updating provides a dynamic retrieval space.
    \item TR achieves approximately 2x speedup on all sizes of LLMs. It achieves a new SOTA with an improvement greater than 31\% compared to previous train-free approaches and even exceeding a training approach.
\end{itemize}





\section{Background}
\label{sec:background}
In this section, we overview the speculative decoding.
We first define auto-regressive (AR) decoding formally, then discuss speculative decoding, focusing on two key strategies: guess-and-verify and tree attention.

\subsection{Auto-Regressive Decoding}
AR is the default decoding strategy of LLMs. 
At each step $t$, LLMs calculate the probability distribution of the next token given the current content $s=(x_0, x_1, \cdots, x_t)$ which $x_i \in \mathcal{V}$:
$$p_{t+1} = P(x| s;\theta).$$
Here $\mathcal{V}$ is the vocabulary and $\theta$ denotes LLM parameters. The next token is selected from $p_{t+1}$ based on the sampling method. Followed~\citet{kou2024CLLMsConsistencyLarge}, we focus on greedy decoding in this paper, where the next token is:
$$x_{t+1}=\text{argmax}\ p_{t+1}.$$
Candidate tokens are the top-$k$ tokens with the highest probabilities
$$
(x_{t+1}^0, x_{t+1}^1, \ldots, x_{t+1}^{k-1}) = \text{argtop}k(p_{t+1})
$$
where $k$ is the number of candidate tokens, $\text{argtop}k(\cdot)$ returns the indices of the top-$k$ highest values in $p_{t+1}$ and $x_{t+1}^0=x_{t+1}$.


\subsection{Speculative Decoding}
\paragraph{Guess and Verify}
Speculative decoding effectively utilizes the parallel capability of accelerators.
Given $s$,
it first guesses $n$ subsequent draft tokens $(\tilde{x}_{t+1},\cdots,\tilde{x}_{t+n})$. 
The combination 
$(s, \tilde{x}_{t+1}, \cdots, \tilde{x}_{t+n})$ is then sent to LLMs for \textbf{one} forward pass, resulting in: 
\begin{align*}
p_{t+1} &= P(x \mid s; \theta), \\
\tilde{p}_{t+i} &= P(x \mid s, \tilde{x}_{t+1}, \ldots, \tilde{x}_{t+i-1}; \theta), i = 2, \ldots, n.
\end{align*}
$p_{t+1}$ is the same as AR decoding so the ground truth $x_{t+1}$ is determinable. 
If the draft token $\tilde{x}_{t+1}$ matches $x_{t+1}$, then $\tilde{p}_{t+2}$ is assumed to identical to $p_{t+2}$. 
Thus, the next ground truth is selected:
$x_{t+2}=\text{argmax}\ \tilde{p}_{t+2}$.
This verification process continues until the draft
token does not match the ground truth, indicated by:
$$x_{t+j} = \text{argmax}\ \tilde{p}_{t+j} \neq \tilde{x}_{t+j}.$$
Ultimately, $j$ new tokens are confirmed in one forward pass. 
The time cost of one forward pass with $(s, \tilde{x}_{t+1}, \cdots, \tilde{x}_{t+n})$ is nearly the same as with $s$ due to the high parallel performance of accelerators. 
Figure~\ref{fig:intro} shows an example. 
The draft tokens are [`i', `in', `range', `('] and the output tokens are [`i', `in', `zip', `(', `xs'] after the forward pass. Though `zip' fails to match `range', three tokens [`i', `in', `zip'] are confirmed in one forward pass. 


\begin{figure*}[t]
\centering
\includegraphics[width=\textwidth]{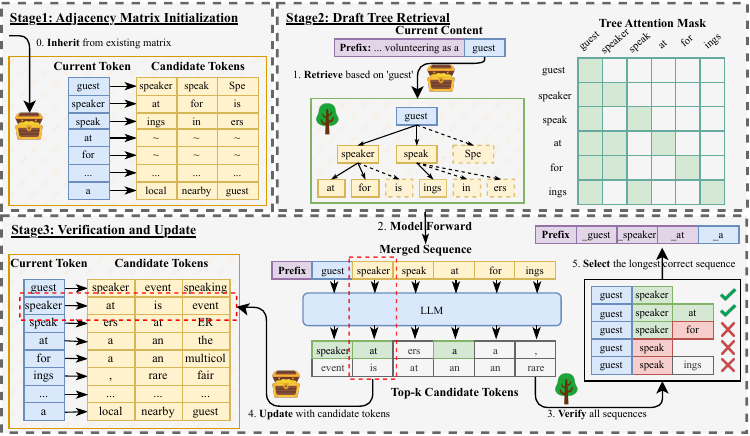} 
\caption{An overview of Token Recycling (TR). The adjacency matrix, initialized by the existing matrix, stores candidate tokens. TR first retrieves a draft tree from the matrix which is then verified through tree attention.
%
After add the longest correct sequence to the content, 
the new top-k candidate tokens update the matrix.
}
\label{fig:method}
\end{figure*}

\paragraph{Tree Attention}
\label{sec:tree_attention}
Traditional causal attention masks are designed for linear sequences, which restricts speculative decoding to verifying one sequence at a time.
However, as the sequence lengthens during draft token generation, the number of potential continuations increases.
For example, in the draft tree in Figure~\ref{fig:method}, the token following `guest' could be `speaker' or `speak'.
Tree attention modifies the attention mask to verify multiple draft sequences simultaneously.
It compresses multiple sequences into a single merged sequence, such as [`guest', `speaker', `speak'], while preserving the tree structure through tree attention mask. Each child node attends only to its parent nodes, preventing sibling tokens from interfering with each other. 
After the LLM processes the merged sequence, all possible sequences such as `guest speaker' and `guest speak', along with their corresponding output tokens are extracted based on the tree structure and verified in parallel.
The longest correct sequence is selected as the final output.
In rare cases, when tokens have identical probabilities, tree attention and AR decoding may select different tokens, but this affects the response quality minimally. The detailed explanation is in Appendix~\ref{sec:same_token_prob}.

In summary, speculative decoding, through \textit{guess and verify} and \textit{tree attention}, improves the inference latency robustly and efficiently.

\section{Methodology}
\label{sec:Methodology}
Figure~\ref{fig:method} provides an overview of Token Recycling (TR).
It leverages a hot-start adjacency matrix to store candidate tokens and employs a BFS-like algorithm to construct a draft tree. 
It utilizes tree attention to verify draft sequences and continuously updates the matrix with new candidate tokens generated during the decoding process. 

\subsection{Adjacency Matrix Initialization}
\label{sec:adj_matrix}

The adjacency matrix $\mathcal{M}$ is a key component in TR, used to store top-$k$ candidate tokens for each token in the vocabulary:
\begin{equation*}
    \mathcal{M} \in \mathcal{V}^{|\mathcal{V}| \times k}
\end{equation*}
where $k$ is a user-defined hyperparameter.
Each element $\mathcal{M}[i, j]$ indicates that the token $V_{M[i,j]}$ is the $j$-th candidate token associated with $V_i$.
The use of matrix format, as opposed to other structures like tries, enables efficient parallel processing of candidate tokens, which is crucial for reducing retrieval and update times. 

Initially, all elements are set to zero, meaning that a token must appear in draft tokens before it has valid candidate tokens.
This initialization leads to the matrix starting with limited predictive capability, potentially causing inefficiencies during the early stages of inference.
To mitigate this limitation, 
we implement a \textit{hot start} strategy. This involves continuing to use the existing matrix, thereby leveraging prior knowledge. Even if queries differ in the domain, candidate tokens often include common expressions and patterns that frequently appear across various queries. 
Consequently, \textit{hot start} ensures that the matrix has a broader starting point, covering a wide range of potential continuations.




\subsection{Draft Tree Retrieval}
\label{sec:retrieval_tree}
The adjacency matrix $\mathcal{M}$ stores candidate tokens, which can be used as draft tokens when their corresponding tokens appear later. 
Directly using the matrix could only determine the immediate next token, such as finding `speaker' following `guest' (see Figure~\ref{fig:method}). Even if `speaker' is correct, it only slightly improves upon AR decoding, adding just one additional token.
In fact, the matrix also holds possible continuations for these candidate tokens, suggesting subsequent tokens like `at' following `speaker'.
Extending the sequence step by step allows for longer draft sequences. Furthermore, by storing top-$k$ candidate tokens, multiple potential continuations can be explored in parallel for each token, such as `at' and `for' following `speaker'.
This BFS process enables the construction of a draft tree with only the adjacency matrix, which can be directly applied to tree attention.

Unlike a complete BFS, we use heuristic rules to define a static and imbalanced tree structure. This tree structure and its construction process are detailed in the Appendix~\ref{sec:tree}. 
\textbf{Static}: The number of children for each node remains constant across all decoding steps, which facilitates pre-processing and enables efficient parallel operations during layer traversal. Avoiding the need to traverse each node individually significantly reduces retrieval time.
\textbf{Imbalance}: Nodes positioned earlier in each layer have more children and extend deeper. This allocates computational resources to the most probable continuations since candidate tokens are ordered by probabilities in the matrix.

The BFS-like approach for retrieving the draft tree begins with the matrix $\mathcal{M}$ and the tree structure $Tree$. 
The root is the last token of current content, like `guest' in Figure~\ref{fig:method}. 
As the root forms the first layer, all candidate tokens for `guest' are extracted from $\mathcal{M}$, resulting in [`speaker', `speak', `Spe']. According to $Tree$, 
the first layer allows each token to have two children,
Therefore, `speaker' and `speak', which have the top-2 probabilities, are added to the second layer. 
The process then proceeds to expand a new layer. 
All candidate tokens of the second layer are retrieved in parallel, resulting in [`at', `for', `is'] and [`ings', `in', `ers']. 
$Tree$ specifies that the first node (`speaker') can have two children, while the subsequent node (`speak') can only have one child. 
Consequently, the new layer tokens are [`at', `for'], and [`ings']. 
This process repeats until the specified depth is reached. 
The detailed Algorithm~\ref{algo:bfs} is provided in Appendix~\ref{sec:tree}.

This retrieval method constructs a draft tree effectively and efficiently with the desired length and variety, which can later be verified by tree attention.

\subsection{Verification and Update}
The verification of the draft tree aligns with Section~\ref{sec:tree_attention}. Merged sequence $S$ is
constructed through traversing the draft tree by layers. All potential draft sequences are then verified and the longest correct sequence is selected.

Following verification, the adjacency matrix $\mathcal{M}$ is updated in parallel based on the output distributions $\tilde{p}_{i+1}$ of each draft token $x_i \in S$: 
\begin{equation*}
\mathcal{M}[\tilde{x}_{i}] = \text{argtop}k(\tilde{p}_{i+1}).
\end{equation*}
Since multiple preceding tokens may have the same candidate token, duplicates may appear in $S$, and their output distributions are likely to differ. 
When performing updates in parallel, CUDA operations may merge these updates, leading to variations in the final result.
For example, if $x_i$ appears twice and has two different top-2 output tokens, $[y_0, y_1], [z_0, z_1]$, then $\mathcal{M}[x_i]$ could be updated to exactly one of the following results: $[y_0, z_1]$, $[y_0, y_1]$, $[z_0, z_1]$ or $[z_0, y_1]$.
We do not resolve this merging, as adding controls reduces overall performance, as discussed later in Section~\ref{sec:update}.

The update process directly overwrites the previous candidate tokens and leverages the new ones as draft tokens for subsequent decoding steps. This allows the retrieval space to dynamically adapt to the current content, focusing on the most relevant and probable continuations.
It also eliminates the necessity for extra operations beyond the standard decoding to update the retrieval space.



\definecolor{mygray}{HTML}{F0F0F0}

\definecolor{mygreen}{HTML}{E3F7E3}


\begin{table*}[ht]
\centering
\small
\begin{tabular}{cc|cccccc|ccc|ccc}
\toprule
\multirow{2}{*}{\#Para} & \multirow{2}{*}{Method} & \multicolumn{9}{c|}{SpecBench} & \multicolumn{3}{c}{MBPP} \\ \cmidrule{3-11} \cmidrule{12-14}
 &  & MT & Trans & Sum & QA & Math & RAG & MAT & Ts/s & Speed & MAT & Ts/s & Speed \\
 \midrule
\multirow{6}{*}{7B} 
~ & AR & 1.00 & 1.00 & 1.00 & 1.00 & 1.00 & 1.00 & 1.00 & 54.30 & 1.00 & 1.00 & 56.15 & 1.00 \\ 
~ & Lade & 1.42 & 1.12 & 1.21 & 1.21 & 1.52 & 1.13 & 1.64 & 69.03 & 1.27 & 1.66 & 79.16 & 1.41 \\ 
~ & PLD & 1.53 & 0.98 & \textbf{2.36} & 1.10 & 1.50 & 1.74 & 1.75 & 83.30 & 1.53 & 1.39 & 66.65 & 1.19 \\ 
~ & REST & 1.37 & 1.05 & 1.12 & 1.42 & 1.06 & 1.30 & 1.84 & 66.29 & 1.22 & 2.08 & 87.08 & 1.55 \\ 
\rowcolor[RGB]{230, 230, 230} \cellcolor{white} ~ & Medusa & 1.90 & 1.57 & 1.48 & 1.58 & 1.87 & 1.45 & 2.31 & 89.41 & 1.65 & - & - & - \\ 
~ & TR & \textbf{2.17} & \textbf{1.90} & 1.94 & \textbf{1.95} & \textbf{2.40} & \textbf{1.78} & \textbf{2.70} & \textbf{110.06} & \textbf{2.03} & \textbf{2.93} & \textbf{131.20} & \textbf{2.34} \\ 
 \midrule
\multirow{6}{*}{13B} 
~ & AR & 1.00 & 1.00 & 1.00 & 1.00 & 1.00 & 1.00 & 1.00 & 39.41 & 1.00 & 1.00 & 41.31 & 1.00 \\ 
~ & Lade & 1.29 & 1.06 & 1.16 & 1.12 & 1.48 & 1.09 & 1.63 & 47.50 & 1.21 & 1.73 & 56.87 & 1.38 \\ 
~ & PLD & 1.45 & 1.01 & \textbf{2.10} & 1.02 & 1.55 & 1.65 & 1.67 & 57.01 & 1.45 & 1.48 & 52.20 & 1.26 \\ 
~ & REST & 1.51 & 1.14 & 1.31 & 1.50 & 1.17 & 1.50 & 1.82 & 53.34 & 1.35 & 2.05 & 70.13 & 1.70 \\ 
\rowcolor[RGB]{230, 230, 230} \cellcolor{white} ~ & Medusa & 1.94 & 1.66 & 1.57 & 1.62 & 1.98 & 1.53 & 2.39 & 67.92 & 1.72 & - & - & - \\ 
~ & TR & \textbf{1.98} & \textbf{1.77} & 1.89 & \textbf{1.75} & \textbf{2.21} & \textbf{1.73} & \textbf{2.72} & \textbf{74.57} & \textbf{1.89} & \textbf{3.08} & \textbf{93.42} & \textbf{2.26} \\ 
 \midrule
\multirow{6}{*}{33B} 
~ & AR & 1.00 & 1.00 & 1.00 & 1.00 & 1.00 & 1.00 & 1.00 & 18.44 & 1.00 & 1.00 & 19.44 & 1.00 \\ 
~ & Lade & 1.32 & 1.09 & 1.20 & 1.17 & 1.55 & 1.14 & 1.61 & 23.03 & 1.25 & 1.70 & 29.22 & 1.50 \\ 
~ & PLD & 1.43 & 1.06 & \textbf{1.94} & 1.08 & 1.55 & 1.41 & 1.55 & 25.89 & 1.40 & 1.41 & 25.89 & 1.33 \\ 
~ & REST & 1.63 & 1.27 & 1.42 & 1.61 & 1.29 & 1.57 & 1.81 & 26.99 & 1.46 & 2.10 & 36.85 & 1.90 \\ 
\rowcolor[RGB]{230, 230, 230} \cellcolor{white} ~ &  Medusa & \textbf{1.98} & \textbf{1.75} & 1.63 & 1.68 & 2.09 & 1.61 & 2.32 & 33.11 & 1.80 & - & - & - \\ 
~ & TR & 1.95 & \textbf{1.75} & 1.92 & \textbf{1.77} & \textbf{2.24} & \textbf{1.78} & \textbf{2.63} & \textbf{35.16} & \textbf{1.91} & \textbf{3.05} & \textbf{45.43} & \textbf{2.34} \\ 
 \bottomrule
\end{tabular}

\caption{Performance of different methods on SpecBench (Vicuna) and on MBPP (Code Llama) across all parameter sizes. 
Speed is the displayed metric for categories of SpecBench.
MBPP results exclude Medusa as it lacks a Code Llama variant.
\colorbox[RGB]{230, 230, 230}{Medusa} involves training while others are training-free. 
\textbf{Bold} represents the highest performance.
}
\label{tab:main_results}
\end{table*}

In summary, TR capitalizes on the `trash' present in speculative decoding by implementing a cycling process between candidate and draft tokens. It accelerates inference without the need for additional model structures or training, making it highly adaptable and seamlessly integrated with any architecture or model size.


\section{Experiment}
\label{sec:experiment}
\subsection{Experimental Setup}
Align with previous work ~\cite{kou2024CLLMsConsistencyLarge}, we focus on common computational redundancy scenarios, specifically greedy decoding with a batch size of one. 
The following evaluation metrics are used: \textbf{Mean Accepted Token (MAT)}~\cite{xia2024UnlockingEfficiencyLarge} represents the average number of tokens confirmed in a single decoding step; \textbf{Tokens per Second (Ts/s)} measures the number of tokens processed per second; \textbf{Speed}up ratio compares the performance relative to HuggingFace's implementation of AR decoding.
We set $k=8$ for $\mathcal{M}$ (\textless2MB storage in sum) and the draft tree structure is shown in Appendix~\ref{sec:tree}.
All experiments are conducted using Pytorch 2.3 with a single A100-80GB GPU and 128 CPUs under CUDA 12.2.

\paragraph{Datasets and LLMs}
\label{sec:datasets}
We conduct experiments on SpecBench~\cite{xia2024UnlockingEfficiencyLarge} and MBPP~\cite{austin2021program}. SpecBench is a comprehensive benchmark 
encompassing diverse scenarios including 
Multi-turn Conversation (MT), Translation (Trans), Summarization (Sum), Question Answering (QA), Mathematical Reasoning (Math), and Retrieval-Augmented Generation (RAG).
MBPP is a widely used dataset in code generation, which has a growing demand for efficient generation.
These datasets enable a comparative analysis with prior work across both general and specialized domains.
We follow the standard practice of utilizing Vicuna~\cite{vicuna2023} for SpecBench and Code Llama~\cite{roziere2023code} for MBPP across three different scales: 7B, 13B, and 33B\footnote{The largest model of Code Llama is 34B, for consistency and convenience in our comparisons, we refer to it as 33B.}.

\paragraph{Baseline}

We compare TR with three train-free retrieval-based methods. 
\textbf{Lookahead (Lade)} constructs an n-gram retrieval library through additional n-gram generation during decoding, consuming significant computational resources. 
\textbf{PLD} treats previous content as the retrieval library, 
which is constrained and cannot introduce new tokens or new token combinations.
\textbf{REST} builds the retrieval library from existing training datasets, requiring large storage and considerable retrieval time. The static nature of the library also prevents it from adapting to individual queries.
Furthermore, we also include a train-need baseline for border comparison.
\textbf{Medusa} adds multiple additional LM heads in the final layer to predict draft tokens. We focus on losses Medusa-1 since Medusa-2 is lossy.
All baselines use their default hyperparameters.

\subsection{Main Results}
\label{sec:main_res}


Table~\ref{tab:main_results} shows the performance of TR compared to other methods. 
On SpecBench, it achieves more than a 2x speedup on the 7B model, nearly 30\% higher than the previous train-free methods. Even compared to tuning Medusa, it shows an improvement of almost 25\%. For the 13B and 33B models, it consistently provides nearly 2x speedup, maintaining the 30\% acceleration advantage. These results demonstrate that TR is the most effective train-free method on SpecBench, offering substantial and consistent speedup across all model sizes.

Notably, TR achieves the best speedup across most sub-tasks as well, except it slightly trails PLD on Sum. This may be due to this task often involves many repetitions of previous content. However, the performance gap between TR and PLD narrows as the model size increases, reaching only a 1\% difference with the 33B model. This is due to larger models tending to generate new tokens rather than repeat previous content.
In other tasks such as MT, Trans, QA, and Math, TR shows a significant improvement of about 40\%\textasciitilde70\% for the 7B model.
This demonstrates the strong generalization of our method across various scenarios. 
Although the improvement on RAG is less than 3\% for the 7B model, it increases with model size, exceeding 10\% for the 33B one.
This improvement is consistent with the preference of larger models for new tokens. 
Compared to the general domain, all methods achieve greater acceleration on the code domain due to its higher content redundancy. TR provides approximately 2.3x speedup across all model scales, achieving the SOTA performance. 

\begin{figure*}[t]
\centering
\includegraphics[width=\textwidth]{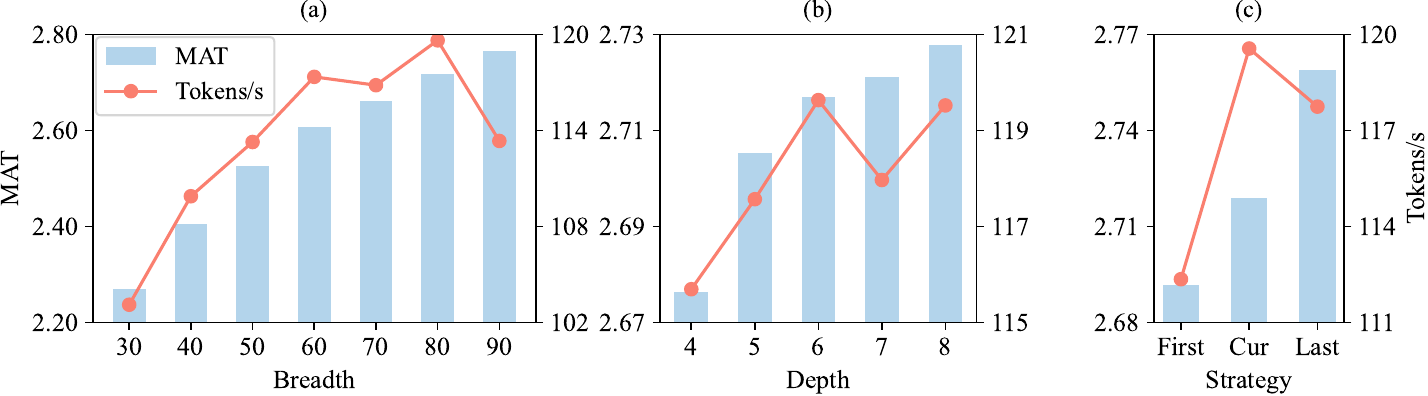} 
\caption{Effects of tree breadth, depth and updating strategies on MAT and Tokens/s are in (a), (b), and (c).}
\label{fig:tree_structure}
\end{figure*}

Furthermore, performances on Trans show the advantages of our method compared to PLD and REST. While PLD shows negligible speedup (close to 1x) and REST achieves its lowest speedup across tasks, TR consistently delivers over 1.75x speedup across all model sizes. Notably, on the 7B model, PLD results in a slowdown, and REST achieves just 1.05x, whereas TR reaches 1.9x.
Trans requires generating new tokens continuously, involving minimal repetition of previous content. Additionally, it is highly context-sensitive, making it challenging to find exact matches from any pre-existing database. These pose challenges for PLD and REST.
In contrast, the adaptive and diverse retrieval space of TR leads to superior performance.
In addition to Speed, TR achieves the highest MAT across both benchmarks. This is attributed to its shorter retrieval times and the avoidance of additional generations like Lade. This allows for deeper and wider draft trees, enabling more tokens to be accepted in a single decoding step. 


\begin{table}[]
\centering
\small
\begin{tabular}{ccc}
\toprule
Method & Memory (MB) & Speed \\
\midrule
Lade & 105 & 1.27 \\
PLD & 0 & 1.53 \\
REST & 465 & 1.22 \\
Medusa & >800 & 1.65 \\
TR & \textbf{1.95} & \textbf{2.03} \\
\bottomrule
\end{tabular}
\caption{The additional memory costs for all methods. Medusa adds extra LM heads to the model, so the memory usage depends on the hidden size and the precision. 800MB is based on a 7B LLM and fp16 precision.}
\label{tab:memory}
\end{table}

Table~\ref{tab:memory} summarizes the GPU memory requirement for all methods. Compared to REST and Lade, TR achieves higher speedup with far less memory. While PLD requires no additional memory, its speedup is limited. Unlike Medusa, our approach is training-free, requires minimal memory, and still achieves superior performance.

TR demonstrates significant improvements across all scenarios, highlighting its efficiency and broad applicability. Importantly, \textbf{TR is train-free and self-drafting, allowing for an approximate 2x speedup that can be seamlessly applied as a `free lunch' to any existing LLM}. 

\section{Analysis}
\subsection{Tree Structure}
\label{sec:tree_structure}
As previously outlined in Section~\ref{sec:retrieval_tree}, our tree structure is static and imbalanced. 
The tree size is a crucial factor to accelerate. 
A larger tree allows more tokens confirmed in one decoding step but also introduces more computational overhead, increasing the time required for each decoding step. To investigate the impact of tree size, specifically its depth and breadth, experiments are conducted on MT-Bench using Vicuna-7B.

\paragraph{Breadth} 
Increasing the breadth of the tree allows for covering more possibilities. In Figure~\ref{fig:tree_structure}(a), the breadth is expanded by adding nodes while keeping the depth fixed at six layers. This leads to a consistent improvement in MAT.
However, when the breadth exceeds 80, Tokens/s begins to decrease. The additional computational overhead eventually outweighs the benefits of a higher MAT.

\paragraph{Depth}
Increasing the depth of the tree allows for accepting longer sequences during decoding. In Figure~\ref{fig:tree_structure}(b), with the number of nodes fixed at 80, the depth is gradually increased.
MAT initially rises rapidly but eventually shows minimal improvement, while Tokens/s noticeably fluctuates. 
Because the matrix stores candidate tokens for only adjacent steps,
longer sequences weaken the connections between distant tokens.
This limitation reduces the effectiveness of increased depth, causing Tokens/s to fluctuate.

\subsection{Ablation Study}

\begin{table}[]
\centering
\small
\begin{tabular}{ccc}
\toprule
~ & Tokens/s & Speed \\
\midrule
AR & 54.98 & 1.00 \\ 
Random & 95.07 & 1.73 \\
Zero & 102.68 & 1.87 \\
Fixed & 117.43 & 2.12 \\
Shuffle & 118.78 & 2.16 \\
TR & \textbf{119.56} & \textbf{2.17} \\
\bottomrule
\end{tabular}
\caption{The impact of different initialization strategies of the adjacency matrix. Random means randomly selected from the vocabulary, Zero means all set to zero, Fixed means inherited from a fixed matrix and Shuffle means shuffle the test set.}
\label{tab:hot_start}
\end{table}

\paragraph{Hot Start}

In TR, the adjacency matrix inherits from the previous one.  
In Table~\ref{tab:hot_start}, we explore the impact of different initialization strategies. 
Random means randomly selecting tokens from the vocabulary, while Zero sets all matrix elements to zero. 
Fixed selects 100 queries from AlpacaEval~\cite{alpaca_eval} (unrelated to the test set), executes them, and stores the resulting matrix. This matrix is then used to initialize each query in the test set.
Shuffle refers to shuffling the test set.
Compared to the Zero, the irrelevant noise introduced by Random leads to a sharp decrease in performance. Fixed, Shuffle and TR show significant improvements over Zero, suggesting that the prior matrix may capture common patterns that effectively assist subsequent queries. The relatively small difference among them indicates that these patterns are generalizable and not tied to specific tasks or content.


\paragraph{Update Strategies}
\label{sec:update}

Section~\ref{sec:adj_matrix} discusses duplicate tokens in the merged sequence during matrix updates. We compare three updating strategies: using candidate tokens from the first occurrence, from the last occurrence, and the current method (merging via parallel CUDA operations). A detailed explanation of three strategies is in Appendix~\ref{sec:explain}
Figure~\ref{fig:tree_structure}(c) indicates that using the last occurrence yields the highest MAT, 
which may benefit from more contextual information. 
However, the differences among different strategies in MAT are minimal.
In terms of Tokens/s, the current approach significantly outperforms the other two, 
as it avoids the additional processing required to manage token positions, thereby reducing delays.
Speculative decoding is highly sensitive to latency, any extra operation must provide substantial benefits to outweigh its time cost.

\paragraph{Effect of Rejected Tokens}
During the update, we refresh the candidate tokens for all draft tokens, including both accepted and rejected tokens. To further illustrate the significant effect of trash tokens, we compare two settings: updating only the candidates of accepted tokens versus of all draft tokens.
As shown in Table \ref{tab:effect_uac_tokens}, including candidates of rejected tokens significantly improves the MAT. This indicates that rejected tokens also carry valuable information necessary for subsequent decoding. We include a case study in Appendix~\ref{sec:case_study}.


\begin{table}[t]
    \centering
    \small
    \begin{tabular}{ccc}
    \toprule
     & SpecBench & MBPP \\
    \midrule
    Only Accepted & 1.63 & 1.99 \\
    All Draft &  \textbf{2.69} & \textbf{2.93} \\
    \bottomrule
    \end{tabular}
    \caption{Mean Accepted Token (MAT) for updating candidate tokens from only accepted or all draft tokens.}
    \label{tab:effect_uac_tokens}
\end{table}

\begin{figure}[t]
\centering
\includegraphics[width=\columnwidth]{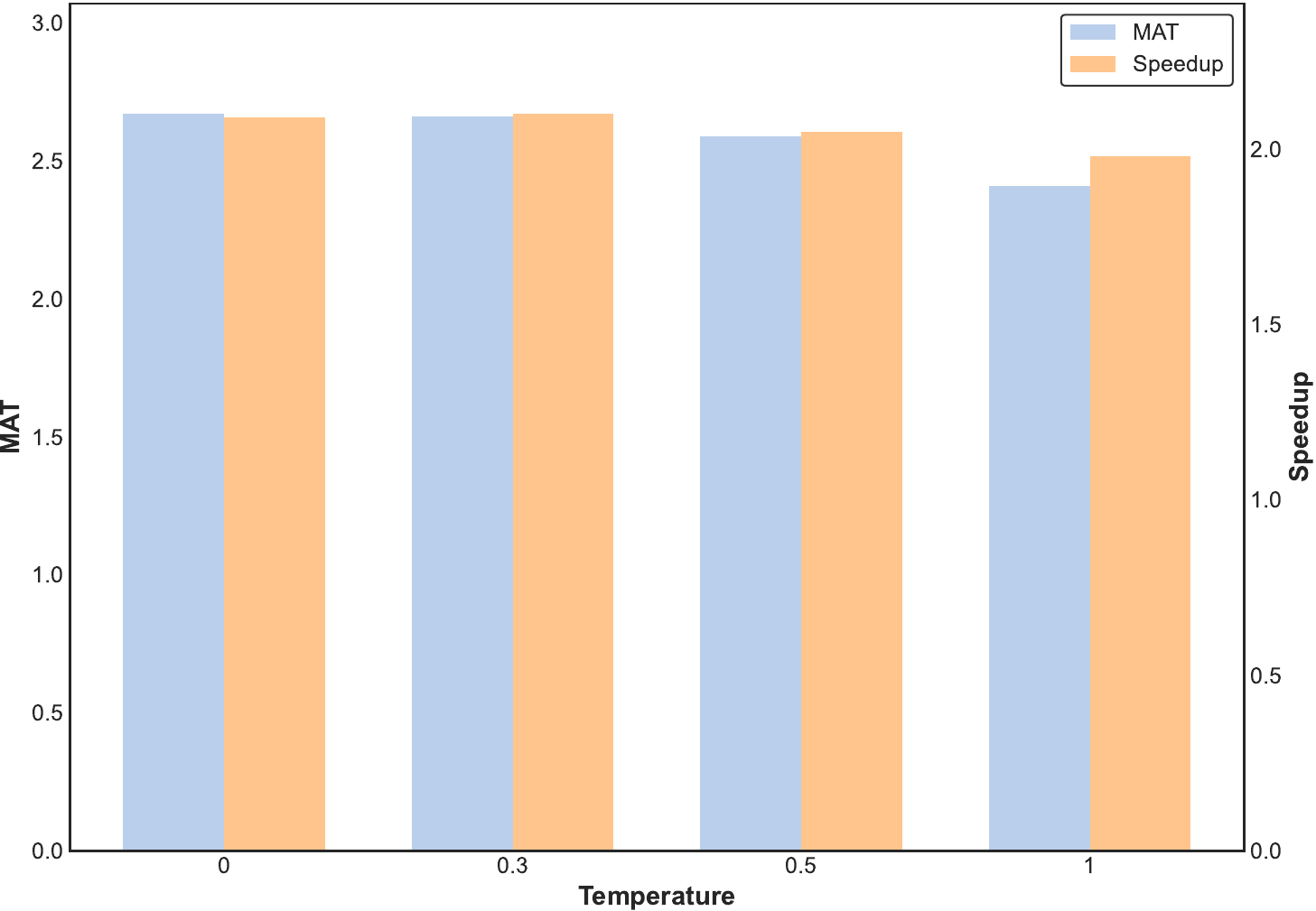} 
\caption{MAT and Speedup ratio under different temperatures during generation.}
\label{fig:temp}
\end{figure}

\paragraph{Temperature Sampling}
To enhance diversity, a temperature greater than 0 is often employed during LLM generation. 
We conducted experiments on SpecBench at different temperature settings to investigate the impact of this randomness on acceleration. 
Figure~\ref{fig:temp} shows that at a temperature of 0.3, the performance remains unaffected when compared to greedy decoding. 
However, as the temperature increases to 0.5 and 1, a slight performance degradation is observed. 
This is likely due to we only store the top-k candidate tokens. 
At higher sampling temperatures, the probability of selecting tokens outside the top-k increases. Despite this, the speedup ratio of TR remains consistently around 2.0. 
The sustained acceleration performance under various sampling settings demonstrates the robustness and broad applicability of TR.

\begin{table}[]
\centering
\small
\begin{tabular}{lccc}
\toprule
Method & Memory (MB) & Tokens/s  & Speed \\
\midrule
Eagle1 & 500 & 106.94 & 2.08 \\
Eagle2 & 500 & \textbf{116.95} & \textbf{2.28} \\
TR & \textbf{1.95} & 107.52 & 2.09 \\
\bottomrule
\end{tabular}
\caption{The comparison of Eagle1/2 with TR about memory costs and speedup ratio.}
\label{tab:eagle}
\end{table}

\subsection{Compare with Eagle}
Table~\ref{tab:main_results} demonstrates that TR significantly outperforms the train-need Medusa. We were curious to compare TR with the state-of-the-art speculative decoding methods - Eagle1~\cite{lieagle} and Eagle2~\cite{li2024eagle}. Eagle1 is a train-dependent method that collects training data and trains a separate draft model using the hidden states and token embeddings from a large LLM. 
Eagle2 improves upon Eagle1 by incorporating dynamic trees into the draft tree construction process.
In Table~\ref{tab:eagle}, TR is compared with Eagle1/2 on SpecBench. It is a surprise that, despite being completely training-free and self-drafting, TR outperforms Eagle1. 
Moreover, TR achieves 91.23\% of the acceleration performance of Eagle2 while only requiring 0.39\% of the memory used by Eagle2.
It is worth noting that TR still employs a static tree, rather than the dynamic tree used in Eagle2. 
This highlights the remarkable effectiveness and efficiency of TR and shows potential for further improvement when combine TR with dynamic trees.

\subsection{Time Allocation}
\label{sec:time}

\begin{figure}[]
\centering
\includegraphics[width=\columnwidth]{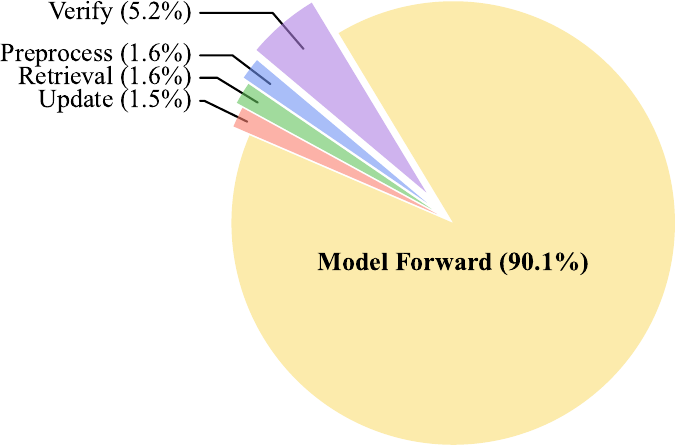} 
\caption{Time allocation for each operation when LLMs respond to a query.}
\label{fig:time}
\end{figure}

Effective speculative decoding requires not only a high hit rate but also minimal time on additional operations.
Each decoding step can be divided into several components: 
\textit{preprocessing} 
, \textit{retrieving} draft tokens, \textit{model forward}, \textit{verifying} draft sequences, and \textit{updating} 
the matrix, input tokens, and key-value cache. 
Figure~\ref{fig:time} shows that the majority of the time is consumed by the model forward. 
The second most is verification which involves extracting and validating all feasible paths.
In contrast, TR introduces only negligible latency in its dedicated preprocessing, retrieval, and update steps. This highlights the efficiency of TR’s design.

\section{Related Work}
\label{sec:related_work}
Efficient inference is crucial for real-time applications and low-resource scenarios.
Many strategies have been developed to reduce latency~\cite{zhou2024survey}. 
Among these, speculative decoding~\cite{chen2023Accelerating,leviathan2023Fast, miao2024SpecInfer, xia-etal-2023-speculative} is a losses technique that predicts multiple possible continuations simultaneously. It reduces the number of decoding steps needed without compromising accuracy. Some speculative decoding methods rely on additional draft models to guess draft tokens. These typically involve using smaller models from the same series~\cite{zhao2024ouroboros, spectoraccelerating, sun2023spectr, pmlr-v235-liu24y, yuan-etal-2024-speculative,gong2024graph} or training new models with a shared vocabulary~\cite{leviathan2023Fast,chen2023Accelerating,zhou2024distillspec,lieagle}. It is worth noting that \citet{zhao2024ouroboros} also uses rejected tokens but does not include candidate tokens.
Additionally, \citet{kou2024CLLMsConsistencyLarge, wang2024make} propose training the original LLMs to enable non-aggressive decoding. While effective, these approaches require managing or training multiple models, which can be non-trivial and resource-intensive.
Other methods focus on parameter-efficient structures. These approaches minimize the need for complete retraining but still require model-specific training and adaptation, limiting their scalability and general applicability~\cite{lin2024BiTA,liu2024Kangaroo}.

Train-free methods construct retrieval libraries to obtain draft tokens~\cite{yang2023inference}. Lookahead~\cite{fu2024BreakSequentialDependency} generates n-grams through multiple decodings, building a retrieval library that can hit multiple tokens in one step. However, it requires the LLM to generate n-grams while responding to queries, which reduces efficiency. PLD~\cite{saxena2023prompt} retrieves only from previous content, resulting in minimal overhead and significant speedup in high-redundancy tasks like summarization. However, it provides little acceleration for tasks requiring the generation of new content, like translation. REST~\cite{he2023rest} constructs retrieval libraries using existing corpora and performs well in common scenarios. However, this approach requires large storage, time-consuming retrieval, and cannot adapt to each query. 

TR is a train-free, retrieval-based method.
It requires no additional generation, covers a broader range of possible continuations, and demands minimal storage with low retrieval costs. The update process ensures an adaptable retrieval space.


\section{Conclusion}

In this work, we introduce Token Recycling, a speculative decoding method for accelerating the inference of LLMs. 
It utilizes an adjacency matrix to store candidate tokens and retrieve a draft tree, which is then verified with tree attention.
The matrix is updated with new candidate tokens generated during decoding.
Token Recycling could be integrated seamlessly with existing LLMs and tasks.
As a train-free approach, 
it achieves a speedup of nearly 2x with \textless2MB of storage, improving over 31\% compared to previous train-free approaches.

\clearpage
\section*{Limitations}
Our study is comprehensive, but has certain limitations that we plan to address in future research. 
In constructing the draft tree, we use a static tree structure. However, a dynamic tree could be employed instead. While dynamic trees introduce additional complexity, they allow for better adaptation to each decoding step, potentially improving performance by tailoring the tree structure to the specific requirements of each query.

\section*{Ethical Considerations}
The data for the proposed methods is drawn solely from publicly accessible project resources on reputable websites, ensuring that no sensitive information is included. 
Moreover, all datasets and baseline models used in our experiments are also available to the public. We have taken care to acknowledge the original authors by properly citing their work.

\section*{Acknowledge}
We gratefully acknowledge the support of the National Natural Science Foundation of China (NSFC) via grant 62236004, 62206078 and 62476073.

\bibliography{custom}

\clearpage
\appendix
\renewcommand{\thesection}{\Alph{section}}
\section{Appendix}
\subsection{Identical Probability Tokens}
\label{sec:same_token_prob}
\begin{table}[h]
    \centering
    \begin{tabular}{ccc}
    \toprule
    Method & MT-Bench & GSM8K \\
    \midrule
    AR Decoding & 6.17 & 35.2 \\
    Tree Attention & 6.23 & 35.2 \\
    \bottomrule
    \end{tabular}
    \caption{Quality/Accuracy comparison of AR-Decoding and Tree Attention on MT-Bench and GSM8K. MT-Bench results are taken from ~\citet{cai2024MedusaSimpleLLM}. It shows that Tree Attention has minimal impact on both answer accuracy and quality.}
    \label{tab:accuracy}
\end{table}

Floating-point representation in the computer has precision errors, commonly known as `floating-point rounding errors'. Specifically, the precision of floating-point numbers is determined by the number of bits in the mantissa. In the IEEE 754 standard, the float32 type has a 23-bit mantissa, meaning the smallest representable difference is $2^{-23}$, approximately $1.19\times10^{-7}$. The float16 type, with a 10-bit mantissa, can represent differences as small as $2^{-10}$, or about $9.77\times10^{-4}$.
If the difference between two token probabilities is smaller than the precision limit of floating-point representation, these two probabilities will be rounded to the same value, and these tokens will be treated as having identical probabilities during sampling.

AR Decoding uses `torch.argmax' to return the token with the highest probability. When the probabilities are the same, `torch.argmax' defaults to returning the one with the smallest index. In Tree Attention, the number of mask tokens is increased compared to AR Decoding, and the attention score of the mask tokens after the softmax operation is not strictly zero, but rather a very small value close to zero. These tiny non-zero values perturb the hidden representations, causing tokens that originally had identical probabilities to now differ slightly, resulting in a different argmax outcome compared to AR Decoding.

Nevertheless, as shown in Table~\ref{tab:accuracy}, due to the extremely rare occurrence of this issue and the affected probabilities being so close to each other, the impact on experimental accuracy and model performance is negligible.

\subsection{Draft Tree Algorithm and Structure}
\begin{algorithm}[t]
\caption{Static Tree Based BFS}
\begin{algorithmic}[1]
\REQUIRE Adjacency matrix $\mathcal{M}$, Static tree structure $Tree$, the last prompt token $x_t$
\ENSURE Merged Sequence $S$
\STATE Initialize $S \leftarrow \emptyset$
\STATE Initialize $root \leftarrow x_t$
\STATE Initialize the current layer $L \leftarrow (root)$
\STATE Initialize the current depth $d  \leftarrow 0$
\WHILE{$d < Tree.depth$}
    \STATE Initialize next layer $L_{\text{next}} \leftarrow \emptyset$
    \STATE Get all candidate tokens of $L$ from $\mathcal{M}$ in parallel
    \STATE $xs = M[L]$
    \STATE Extract next layer tokens from $xs$ with $Tree$
    \STATE $L_{\text{next}} = xs[Tree[d].index]$
    \STATE Concatenate $S$ and $L$
    \STATE $S \leftarrow (S;L)$
    \STATE $L \leftarrow L_{\text{next}}$
\ENDWHILE
\RETURN $S$
\end{algorithmic}
\label{algo:bfs}
\end{algorithm}
\label{sec:tree}
Utilizing tree attention ~\cite{miao2024SpecInfer} to extend the path in the verification phase has become a widely adopted strategy for speculative decoding methods.

\begin{figure*}[t]
\centering
\includegraphics[width=\textwidth]{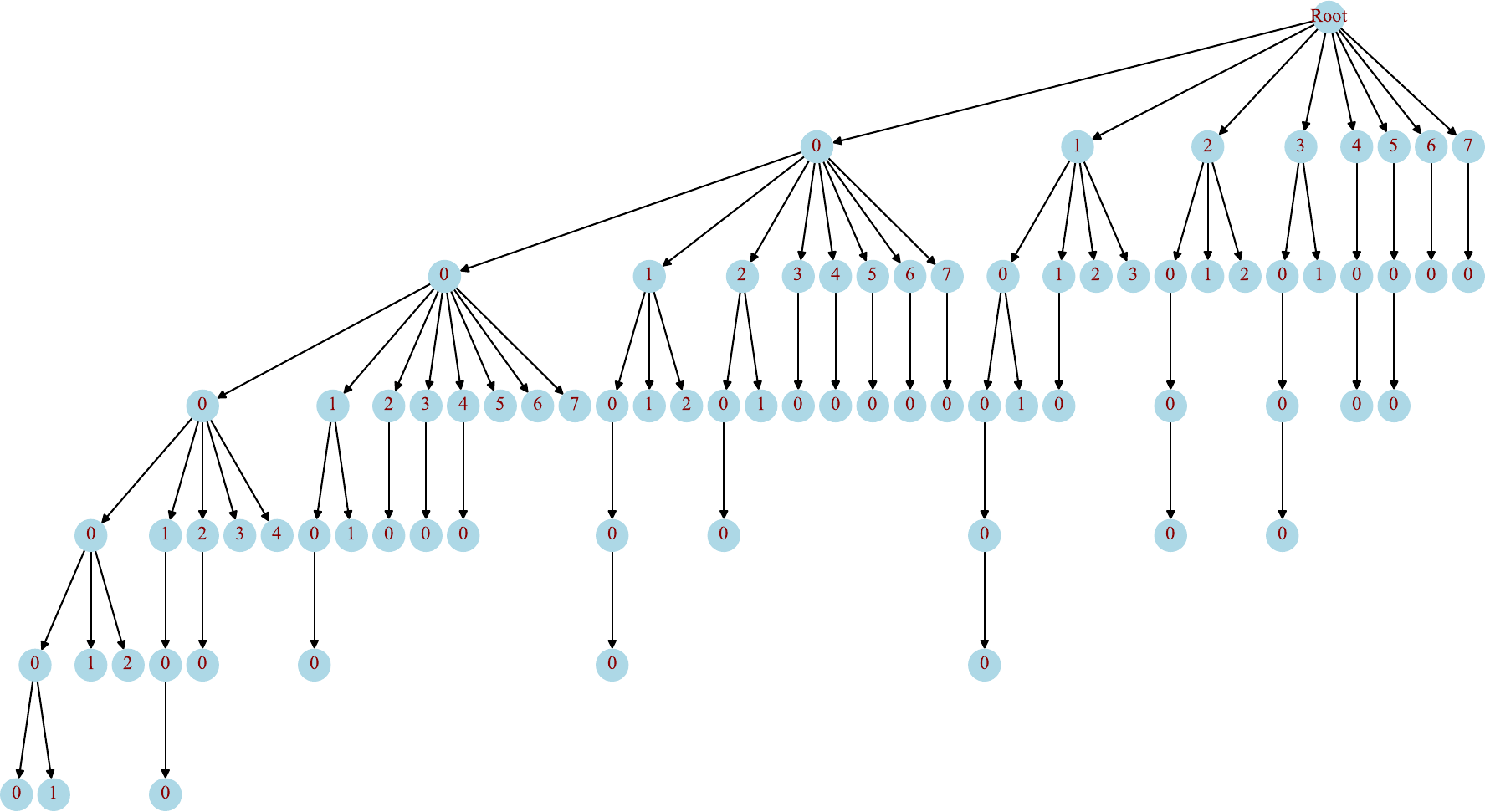} 
\caption{The static tree used in Token Recycling.}
\label{fig:static_tree}
\end{figure*}

In Token Recycling, we also use a heuristically constructed token tree to perform the verification.
As shown in Figure \ref{fig:static_tree}, we construct a static and unbalanced tree inspired by ~\citet{cai2024MedusaSimpleLLM}. The number $k$ on a node indicates that it is the $k$-th candidate token for its parent node.
The construction process is below. We begin with a fully balanced 10-branch tree and use an independent validation set to identify the top $k$ nodes that most frequently yield correct tokens. These top $k$ nodes and their children are retained to form a new tree, and the process is repeated to identify the next set of top $k$ nodes. This iterative process continues until performance no longer shows significant improvement. The final tree is determined, and the $k$ is set to consider the maximum number of children across all nodes and the memory requirement. While empirical, this iterative approach has proven to be effective. Further details on tuning the $n$ are provided in Section~\ref{sec:tree_structure}.
Overall, the tree we construct contains 81 nodes (including the root node) in 6 layers.
This means that each forward requires an additional draft input of 79 tokens with a maximum acceptance length of 6.

Building on the tree structure described above, we construct a draft tree for the current content by a BFS-like algorithm in the inference phase.
As described in Algorithm \ref{algo:bfs}, we infill the child nodes of each layer in turn according to the matrix.
At last, the merged sequence $S$ is returned and sent to tree attention with $Tree$.

\subsection{Reuse Mechanism Analysis}
\label{sec:reuse}
The output of LLM is context-dependent, it makes us curious why reusing candidate tokens from previous generations works. 
\citet{xiao2024efficient} analyzed the distribution of attention logits in Transformers and found that the first two layers focus more on `local' patterns, with recent tokens receiving much more attention. In later layers, the model shifts its focus to the tokens at the beginning of the sequence. To accelerate, candidate tokens need to satisfy both local semantics and long-range dependencies.
As shown in Figure~\ref{fig:intro}, draft tokens are divided into accepted and rejected tokens. Accepted tokens are those that appear in the previous query response. In Token Recycling, candidate tokens for all draft tokens are stored, not only accepted ones. Each decoding step involves 79 draft tokens, meaning 79 tokens receive/update their candidate tokens at each step. The quantity of rejected tokens is far more than accepted tokens, and their candidate tokens are also stored in the matrix. In other words, previous generations actually provide a large number of common patterns stored in the matrix, and these patterns often meet local semantic needs.

From two perspectives, the reuse of these common patterns is justified:

For scenarios like sentence transitions, verb collocations, punctuation, or words split into multiple tokens, there is often no need for long-range dependencies. The common patterns stored in the matrix can significantly accelerate the decoding process.

Rejected tokens may not satisfy the long-range dependencies of the previous, but this does not mean they do not meet the long-range dependencies of the current. These tokens may be accepted in the future. As shown in Table~\ref{tab:effect_uac_tokens}, the performance gain from the candidate tokens of rejected tokens is significantly greater than the gain from using only the accepted tokens in the SpecBench experiments. We also include a case study in \ref{sec:case_study}.

\subsection{Case Study}
\label{sec:case_study}
Below, we present a real example from MT-Bench, illustrating how accepted tokens and rejected tokens contribute to the acceleration of Token Recycling. First Round:
\begin{itemize}
    \item Prefix last token: ['guest']
    \item Merged sequence with draft tokens: ['guest', 'speaker', '</s>', 'speak', '<0x0A>', 'Spe', 'lect', 'speaking', 'spe', 'at', 'for', '</s>', 'is', 'could', 'can', 'would', '<0x0A>', '<s>', 'sime', 'multicol', 'bolds', '</s>', 'ings', 'in', '<0x0A>', 'a', 'aking', 'ures', 'engag', 'aking', 'a', 'an', 'the', '[', 'our', 'local', '</s>', 'up', 'a', 'an', '</s>', '<s>', 'sime', 'a', 'be', 'help', 'like', '<0x0A>', 'The', 'Home', 'guest', '<s>', '<0x0A>', 'opportunity', '</s>', 'local', 'nearby', 'guest', 'up', 'public', 'guest', '</s>', 'guest', 'public', 'local', 'local', 'The', 'first', 'The', '<0x0A>', 'event', 'community', 'Toast', 'Buddh', 'speaker', 'speaker', 'event', 'time', '.', ',', 'ism']
    \item Accepted sequence: ['guest', 'speaker', 'at', 'a', 'local', 'event', 'could']
\end{itemize}

Key observations: 
\begin{itemize}
    \item \textbf{Rejected tokens still receive candidate tokens}: For example, ['be'] was not directly accepted, but its candidate tokens were stored.
    \item \textbf{Candidate tokens of accepted tokens are stored}: ['local'] for ['a'] was chosen in this round, but other candidate tokens were also retained.
\end{itemize}

At this point, the adjacency matrix stores:
\begin{itemize}
    \item 'be': ['able', 'onto', 'mistaken', 'wrong', 'the', 'interested', 'persu', 'a']
    \item 'a': ['local', 'time', 'personal', 'professional', 'few', 'unique', 'great', 'low']
\end{itemize}

Second Round:
\begin{itemize}
    \item Current prefix last token: ['could']
    \item Merged sequence with draft tokens: ['could', 'be', 'provide', '</s>', 'help', 'offer', 'present', 'not', 'actually', 'a', 'an', 'the', '</s>', 'just', 'one', 'benef', 'both', ',', "'", 'you', 'for', '<s>', 'sime', 'multicol', 'you', 'overcome', 'some', '</s>', 'public', '<unk>', 'great', 'fant', 'wonderful', 'valuable', 'unique', 'perfect', '</s>', 'ter', 'up', 'event', 'local', 'up', 'local', '<s>', 'not', 'of', 'ited', 'models', 'such', 'like', 'my', 'The', 'as', 'comp', '<s>', 'opportunity', 'guest', '</s>', 'speaker', 'way', 'astic', 'asy', 'bl', 'ins', 'guest', 'coming', 'coming', 'as', 'first', 'a', 'at', 'is', 'would', 'event', '<s>', 'natural', 'public', 'a', 'a', 'an', 'could']
    \item Accepted sequence: ['could', 'be', 'a', 'great']
\end{itemize}

At this point:
\begin{itemize}
    \item \textbf{Candidate tokens from rejected tokens are matched}: 'be' correctly predicted 'a'.
    \item \textbf{Candidate tokens from accepted tokens are matched}: 'a' successfully predicted 'great'.
\end{itemize}

\subsection{Explanation of Three Update Strategies}
\label{sec:explain}
Suppose our merged sequence is:
$$ S = [\underline{a}, b, c, \underline{a}, d] $$,
which token $a$ appears twice.
Currently, corresponding output tokens are stored in matrix $O$:
$$
\begin{aligned}
O[0] &= [a_1, a_2, a_3, a_4] \\
O[1] &= [b_1, b_2, b_3, b_4] \\
O[2] &= [c_1, c_2, c_3, c_4] \\
O[3] &= [a_5, a_6, a_7, a_8] \text{ (second occurrence of } a \text{)} \\
O[4] &= [d_1, d_2, d_3, d_4]
\end{aligned}
$$

The three strategies for updating the adjacency matrix $M$ are:
\begin{itemize}
    \item Cur (uncontrolled): Update directly without position control, potentially mixing outputs from repeated tokens. For example: $ M[S]=O, M[a] = [a_1, a_2, a_3, \underline{a_8}]$ (the last $\underline{a_8}$ is from the second occurrence).
    \item First: Record positions of each token's first occurrence: $\text{pos} = [0(a), 1(b), 2(c), 4(d)]$. Resulting in: $M[S[pos]]=O[pos], M[a] = [a_1, a_2, a_3, a_4]$
    \item Last: Record positions of each token's last occurrence: $\text{pos} = [1(b), 2(c), 3(a), 4(d)]$. Resulting in: $M[S[pos]]=O[pos], M[a] = [a_5, a_6, a_7, a_8]$
\end{itemize}

These different update strategies have a slight impact on Mean Accepted Tokens. However, extracting these positions introduces additional latency, ultimately reducing overall acceleration effectiveness compared to the simpler uncontrolled approach.

\end{document}